\documentclass[lettersize,journal]{IEEEtran}
\usepackage{amsmath,amsfonts}
\usepackage{algorithmic}
\usepackage{algorithm}
\usepackage{array}
\usepackage[caption=false,font=normalsize,labelfont=sf,textfont=sf]{subfig}
\usepackage{textcomp}
\usepackage{stfloats}
\usepackage{url}
\usepackage{verbatim}
\usepackage{graphicx}
\usepackage{cite}
\usepackage{hyperref}  %hyperref still needs to be put at the end!
\usepackage{bm}
\usepackage{xcolor}
\usepackage{colortbl}
\usepackage{multirow}
 
\usepackage{amsmath}
\usepackage{amsfonts}
\usepackage{amssymb}
\begin{document}

\title{PACF: Prototype Augmented Compact Features \\ for Improving Domain Adaptive Object Detection}

\author{Chenguang Liu$^{\dag}$, Yongchao Feng$^{\dag}$, Yanan Zhang, Qingjie Liu$^{*}$, ~\IEEEmembership{Member,~IEEE},

and Yunhong Wang, ~\IEEEmembership{Fellow,~IEEE}
        % <-this % stops a space
\thanks{
$^{\dag}$These two authors contributed equally to this work.
$^{*}$Corresponding author.

Chenguang Liu, Yongchao Feng, Qingjie Liu, and Yunhong Wang are with the State Key Laboratory of Virtual Reality Technology and Systems, Beihang University, Beijing 100191, China (e-mail: liuchenguang@buaa.edu.cn; fengyongchao@buaa.edu.cn; qingjie.liu@buaa.edu.cn; yhwang@buaa.edu.cn).

Yanan Zhang is with School of Computer Science and Information Engineering, Hefei University of Technology, Hefei 230601, China (e-mail: yananzhang@hfut.edu.cn).

Qingjie Liu, and Yunhong Wang are also with the Hangzhou Innovation Institute, Beihang University, Hangzhou 310051, China.
}}

\markboth{Journal of \LaTeX\ Class Files,~Vol.~14, No.~8, August~2021}%
{Shell \MakeLowercase{\textit{et al.}}: A Sample Article Using IEEEtran.cls for IEEE Journals}


\maketitle

\begin{abstract}
In recent years, there has been significant advancement in object detection.
However, applying off-the-shelf detectors to a new domain leads to significant performance drop, caused by the domain gap. These detectors exhibit higher-variance class-conditional distributions in the target domain than that in the source domain, along with mean shift.
To address this problem, we propose the \textbf{P}rototype \textbf{A}ugmented \textbf{C}ompact \textbf{F}eatures (PACF) framework to regularize the distribution of intra-class features.
Specifically, we provide an in-depth theoretical analysis on the lower bound of the target features-related likelihood and derive the \textit{prototype cross entropy} loss to further calibrate the distribution of target RoI features. 
Furthermore, a \textit{mutual regularization} strategy is designed to enable the linear and prototype-based classifiers to learn from each other, promoting feature compactness while enhancing discriminability.
Thanks to this PACF framework, we have obtained a more compact cross-domain feature space, within which the variance of the target features' class-conditional distributions has significantly decreased, and the class-mean shift between the two domains has also been further reduced. 
The results on different adaptation settings are state-of-the-art, which demonstrate the board applicability and effectiveness of the proposed approach.
\end{abstract}

\begin{IEEEkeywords}
Domain adaptation, Object detection, Prototype, Compact features.
\end{IEEEkeywords}

\section{Introduction}
\label{sec:intro}

\IEEEPARstart{A}{s} a foundational task in computer vision, object detection has experienced rapid development in recent years~\cite{ren2015faster, cai2018cascade, carion2020detr}.
Well-performing detectors typically rely on the assumption that training and testing data are independently and identically distributed.
However, deploying these methods in a novel domain leads to the catastrophic performance degradation due to the domain gap~\cite{chen2018domain}, which significantly limits the generalization and transferability of object detectors.   

\begin{figure}[t]
  \centering
  \includegraphics[width=0.47\textwidth]{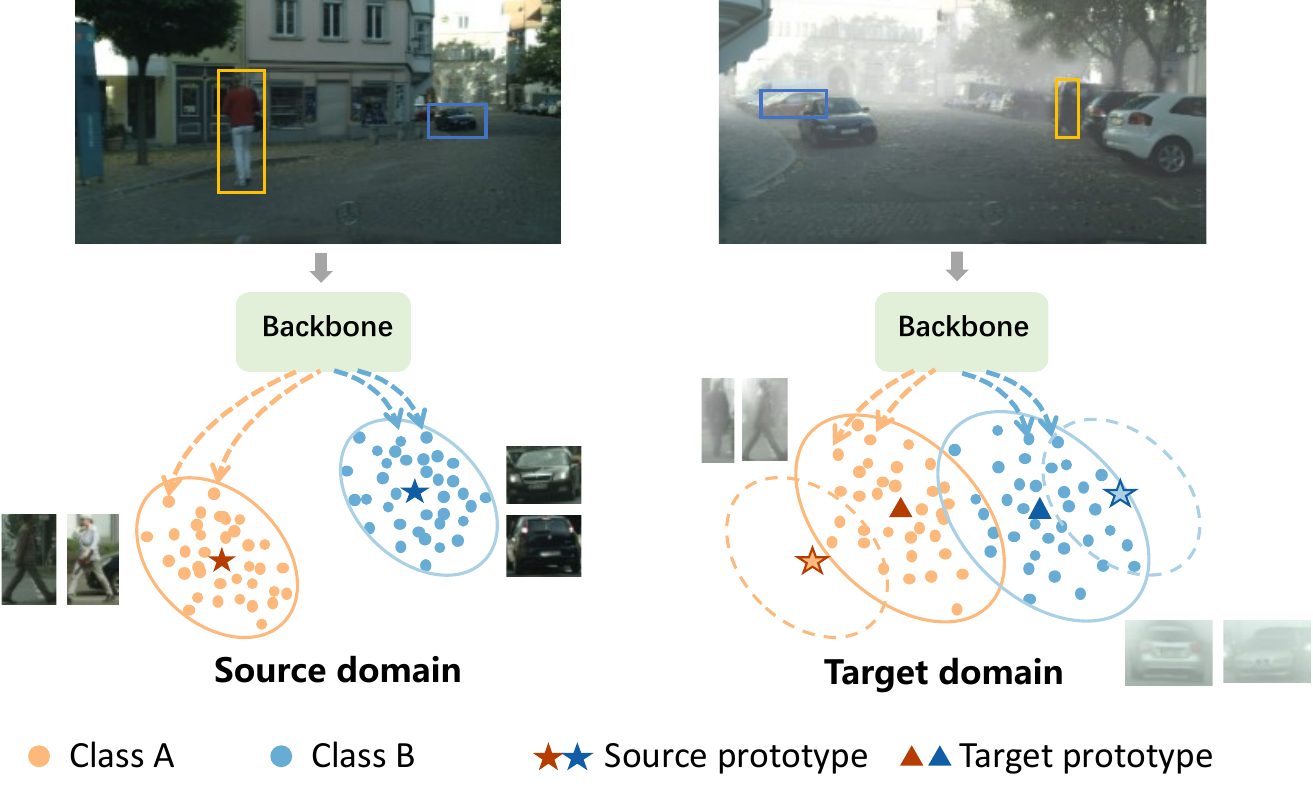}
  \caption{When a detector trained on the source is directly applied to target images, the domain gap causes a shift in feature distribution. We observe that intra-class conditional distribution which is compact on the source tends to scatter on target images, where target feature distribution exhibits mean shift and larger intra-class variance compared with the source.
  }
  \label{fig:problem}
\end{figure}

Domain Adaptive Object Detection (DAOD)~\cite{vu2024supervision, fengdsd, jeon2024raw, wang2024triple, piao2024unsupervised, liu2023foregroundness} have been explored to address this issue. DAOD methods transfer detectors trained on an annotated source domain to an unlabeled target domain. In recent years, researchers have proposed considerable approaches and one of the main streams of DAOD works is to align feature distributions between source and target domains. Early works ~\cite{chen2018domain, saito2019strong, zheng2020cross, zhao2020adaptive} try to extract multi-level domain-invariant features between different domains through designing a domain discriminator to achieve domain alignment. Some works aim to align the cross-domain class-conditional distribution in the feature space and achieve adaptation in a category-wise manner.
Recently, based on teacher-student framework, methods~\cite{li2022cross,cao2023contrastive, chen2022learning, do2024d3t, liu2024decoupled} represented by AT ~\cite{li2022cross} combine adversarial training and self-training~\cite{khodabandeh2019robust, kim2019self} to iterate continuously for improving the detection performance on the target domain and achieve the state-of-the-art performance.

However, these works neglect the significant intra-class variance and directly align category centers, which inevitably result in a suboptimal adaptation. 
In the DAOD task, domain gap is characterized by mean shift and variance changes in the class-conditional distributions from source to target domain. 
As shown in Fig.~\ref{fig:problem}(a), considering domain style transformation (\textit{e.g.} clear $\rightarrow$ foggy), the appearance of the objects may change significantly. These changes include blurriness of object contours, occlusions of salient regions, and more complex background, resulting in the extracted features from the backbone becoming more divergent and exhibiting larger variance.
The intra-class variance contains essential information about class-conditional distributions, which should also be aligned for domain adaptation. Overlooking the intra-class variance could lead to lots of non-adapted objects, as well as the potential overlapping of different class-conditional distributions with classification errors~\cite{li2022sigma}.

When the intra-variance of a certain category becomes large, its distribution tends to scatter rather than be compact on the target domain. This can lead to significant overlap between different classes, particularly for appearance-similar categories. Additionally, the detector tends to find inaccurate decision boundaries, and susceptible to noisy data or outliers.

Because the detector fits the source data during the training process, the likelihood value of source RoI features is higher, reflecting strong adaptation to source data distribution. However, the likelihood of target RoI features tends to be lower due to the domain gap, resulting in suboptimal performance on the target domain. To address this, we aim to bridge the domain gap by maximizing the likelihood of target RoI features as the objective function.

To achieve this, we introduce domain-specific class \textit{prototype} and propose \textit{prototype cross entropy} loss to regularize the feature representation: 
target features could be close to the source and target prototypes belonging to the same class, while staying far from prototypes of other classes, reducing mean shift between the two domains and facilitating the extraction of more compact features.
Specifically, we introduce the source and target domain prototypes for each class. Given target features $x^t$, we attain the similarity between $x^t$ and prototypes from two domains via `cosine-softmax’ operation for classification. The linear classifier has learnable parameters that could be updated through back-propagation, while the prototype-based classifiers rely on domain-specific class prototypes that are updated in the training.

Inspired by these observations, we propose \textbf{P}rototype \textbf{A}ugmented \textbf{C}ompact \textbf{F}eatures (PACF) framework to regularize the distribution of intra-class features via \textit{prototype cross entropy} loss.
However, this loss encourages target domain features to become compact, declining in discriminability—a property that is particularly well-captured by the linear classifier.
Therefore, we impose \textit{mutual regularization} on the two types of classifiers to balance feature compactness and discriminability, thereby further enhancing the model's performance.
The contributions of this paper are listed as follows:

\begin{itemize}
\item We provide an in-depth theoretical analysis on the lower bound of the target features-related likelihood, and derive \textit{prototype cross entropy} loss to empirically optimize this lower bound, leading to a more compact target domain feature space.
\item We propose a \textit{mutual regularization} strategy to promote reciprocal learning between the two types of classifiers, balancing target features' compactness and discriminability.
\item We have conducted extensive experiments under different adaptation settings, and our method outperforms the state-of-the-art by a large margin, \textit{e.g.}, improving performance on Cityscapes$\rightarrow$Foggy Cityscapes from 50.3\% to 52.3\%.
\end{itemize}

\begin{figure*}[t]
  \centering
  \includegraphics[width=0.96\textwidth]{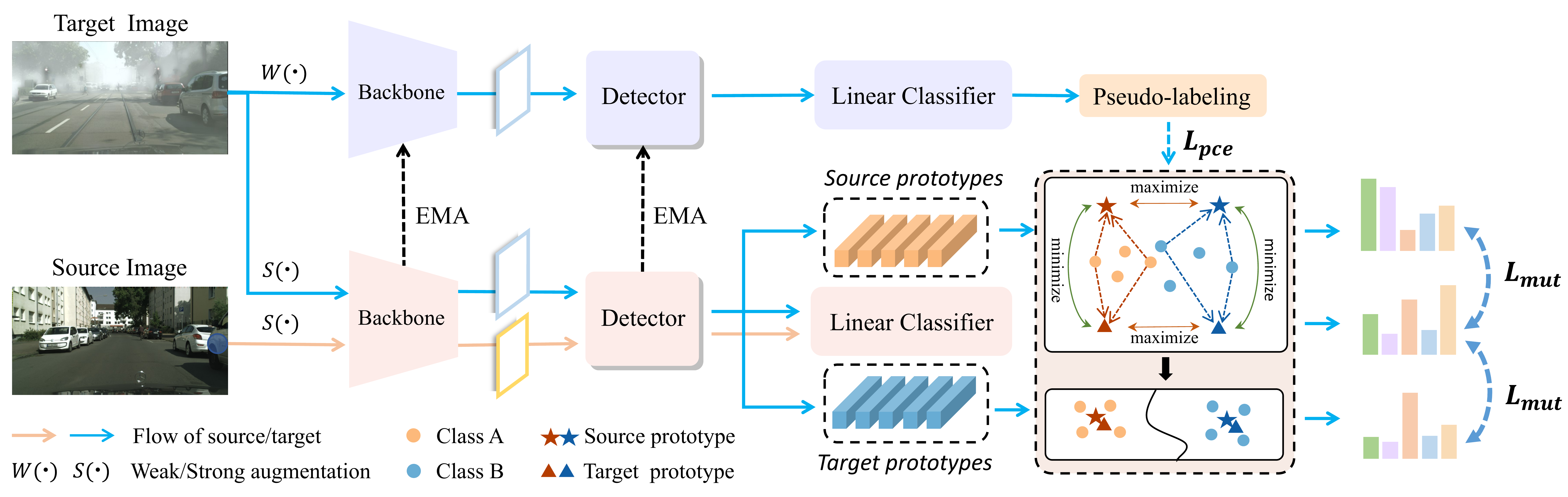}
  \caption{The overview of the proposed Prototype Augmented Compact Features (PACF) framework for DAOD. The teacher model adopts the target images with weak augmentation and the predictions of the teacher model are filtered as pseudo labels. Simultaneously, the source and target images with strong augmentation are fed into the student model. The prototype cross entropy loss $L_{pce}$ encourages target features to simultaneously move closer to the prototypes of both domains belonging to the same class, while staying far from prototypes of other classes. In addition, a mutual regularization loss $L_{mut}$ is proposed to balance feature compactness and discriminability on the target domain.}
  \label{fig:overall}
\end{figure*}

\section{Related Work}

\subsection{Domain Adaptive Object Detection}

Domain adaptive object detection (DAOD) leverages a labeled domain to learn a detector generalizing to a novel domain free of annotations. Recent research in DAOD can also be broadly divided into two primary categories: adversarial training and self-training.

\textbf{Adversarial Training}.
Adversarial training methods~\cite{vs2021mega, zhao2022task} often utilize a domain discriminator to train feature encoder and discriminator in an adversarial manner. DA-Faster~\cite{chen2018domain} first proposes to perform adversarial with a gradient reverse layer (GRL)~\cite{ganin2015unsupervised} at image-level and instance-level in object detection.
SWDA~\cite{saito2019strong} proposes strong-weak distribution alignment, employing strong alignment locally and weak alignment globally. 
Zheng \textit{et al.}~\cite{zheng2020cross} utilize the coarse-to-fine adaptation by applying gradient reversal-based adversarial loss at multiple layers. Zhao \textit{et al.}~\cite{zhao2020adaptive} combine weak global alignment with multi-label classification, using the prediction scores of the multi-label classifier as conditions for the domain discriminator. PARPN~\cite{zhang2021rpn} extends the idea of feature alignment in the RPN stage to narrow down the domain shifts.
MeGA-CDA~\cite{vs2021mega} performs category-aware domain alignment via $K$ category-wise discriminators.
Some methods~\cite{lin2021domain, liu2022decompose} decompose features into domain-invariant and domain-specific features. 
DDF~\cite{liu2022decompose} proposes multiple extractors to obtain domain-shared and domain-private features respectively. TFD~\cite{wang2024triple} designs Multi-level Disentanglement Module to enhance domain-invariant representations, while utilizing Cyclic Disentanglement Module to facilitate domain-specific representations.

\textbf{Self-training}.
Self-training~\cite{roychowdhury2019automatic, zhao2020collaborative} involves generating pseudo labels for target domain data using a pretrained detector on the source domain and subsequently retraining the model with these pseudo labels in the target domain. Roychowdhry~\cite{roychowdhury2019automatic} \textit{et al.} leverages video data from the target domain to automatically compute pseudo labels. D-adapt~\cite{jiang2021decoupled} proposes a decoupled approach to separate adaptation from detection. TDD~\cite{he2022cross} proposes a target proposal perceive module to adaptively guide source detection branch to obtain more accurate target pseudo labels. FA-TDCA~\cite{liu2023foregroundness} employs the foregroundness concept to select high-quality pseudo labels, and adopts a self-paced curriculum learning paradigm to gradually improve the pseudo label qualities of target domain data.

Recently, the mean teacher paradigm has shown promising performance, with the teacher providing pseudo labels, and the student progressively updating the teacher. AT~\cite{li2022cross} combines adversarial training and self-training in a unified framework. CMT~\cite{cao2023contrastive} leverages pseudo label to extract object-level features and optimize them through contrast learning. DSD-DA~\cite{fengdsd} addresses source bias and enhances classification-localization consistency through a distillation-based approach and a domain-aware strategy. 2PCNet~\cite{kennerley20232pcnet} proposes a two-phase strategy to combine high and low confidence pseudo labels with domain specific augmentation. DUT~\cite{liu2024decoupled} proposes cross-domain feature intervention to enhance the target model with better generalization ability and robust detection capacity to unseen environments.
However, these works ignore the significant intra-class variance, leading to a suboptimal alignment of class-conditional distributions. Our work breaks this barrier with prototype cross entropy loss, avoiding the inaccurate adaptation caused by handcraft prototype design and center-based alignment.

\subsection{Prototype for Domain Adaptive Object Detection}

The \textit{prototype} serves as the representative features of a particular class and has been applied in DAOD~\cite{belal2024multi,zhang2021c2fda, zheng2020cross,chen2021i3net,chen2021dual,xu2020cross,zhang2023multi,zhang2021rpn,li2022sigma}. The usage of prototypes in these works typically falls into two categories: 

1) Some methods employ prototypes as representatives of features within the same class, utilizing various losses (\textit{e.g.}, contrastive or L2 loss) to pull prototypes of the same class closer across different domains while separating prototypes of different classes within the same domain. The differences of these works lies in that how to model the prototypes: GPA~\cite{xu2020cross} models instance-level features within a domain as graph structures, extracting the average node features of a certain class instance as prototypes, whereas DBGL~\cite{chen2021dual} computes the average of pixel values within annotated boxes to obtain pixel-level prototypes of a certain class. However, these methods only utilize class prototypes to construct losses lacking the direct involvement of RoI features, failing to adequately regularize feature distributions in the space. In this work, we emphasize the progressive movement of target features toward shared prototypes across domains. Therefore, our work concentrates on theoretical analysis of target features' likelihood, aiming to mitigate mean shift and larger intra-class variance in a unified framework.

2) Other methods consider prototypes as tools to assist domain adaptation. For example, SIGMA~\cite{li2022sigma} treats prototypes as the mean of class-condition Gaussian distributions, sampling points from this distribution for semantic completion. MPG~\cite{zhang2023multi} uses similarity scores between instance features with prototypes to further refine pseudo labels from linear classification results, thereby obtaining more accurate pseudo labels. In this work, we construct a prototype-based classifier to classify RoI features. During this process, prototypes and RoI features directly compute similarity for classification, encouraging RoI features to be closer to the same class prototype and away from other classes' prototypes, resulting in a more compact class-conditional distribution.

\section{Method}

\subsection{Problem Formulation}
In the context of domain adaptive object detection, we are given a labeled source domain $\mathcal{D}_S = \{ (x_i^s, y_i^s) \}_{i=1}^{N_s}$, where $x_i^s$ represents the $i_{th}$ image and $y_i^s=(b_i^s, c_i^s)$ denotes its corresponding labels, consisting of the bounding box coordinates $b$ and category $c$. Additionally, we have an unlabeled target domain $\mathcal{D}_T = \{ x_i^t \}_{i=1}^{N_t}$. The source and target samples are assumed to be drawn from different distributions ($\mathcal{D}_S \neq \mathcal{D}_T$), but share the same categories. The goal is to improve the detection performance in $\mathcal{D}_T$ by leveraging the knowledge from $\mathcal{D}_S$.

\subsection{Framework Overview}
The overview of our framework is presented in Fig.~\ref{fig:overall}, which is build upon representative domain adaptive object detection method AT~\cite{li2022cross}. Our framework consists of target-special teacher model and cross-domain student model, which share the same architecture. In the training, weakly-augmented target images are fed into teacher while strongly-augmented source and target images pass through student. 
Firstly, we employ the discriminator and the gradient reverse layer (GRL)~\cite{ganin2015unsupervised} to align the distributions across two domains in student model.

Furthermore, we introduce the source and target domain prototypes for each class. In the student model, target features move closer to their respective prototypes in a class-specific manner for both domains, while stay from other prototypes. By introducing a regularization term between two types of classifiers, the linear classifier is also encouraged to obtain compact features.

Finally, the student leverages ground truths and pseudo labels provided by the teacher for supervision of source and target images, while the teacher updates its weights through exponential moving average (EMA) of the student. The overall loss of AT~\cite{li2022cross} is formulated as:
\begin{equation}
\begin{aligned}
\label{eq:0}
L_{AT} = L_{sup} + \lambda_{unsup}\cdot L_{unsup} + \lambda_{dis}\cdot L_{dis}
\end{aligned}  
\end{equation}
where $\lambda_{unsup}$ and $\lambda_{dis}$ are the hyper-parameters, we set the $\lambda_{unsup} = 1.0 $ and $\lambda_{dis} = 0.1$ for all the experiments. $L_{sup}$ and $L_{unsup}$ are utilized to train the student model on labeled source and unlabeled target images, respectively, while $L_{dis}$ is employed for adversarial training. 

\begin{figure}[t]
  \centering
  \includegraphics[width=0.50\textwidth]{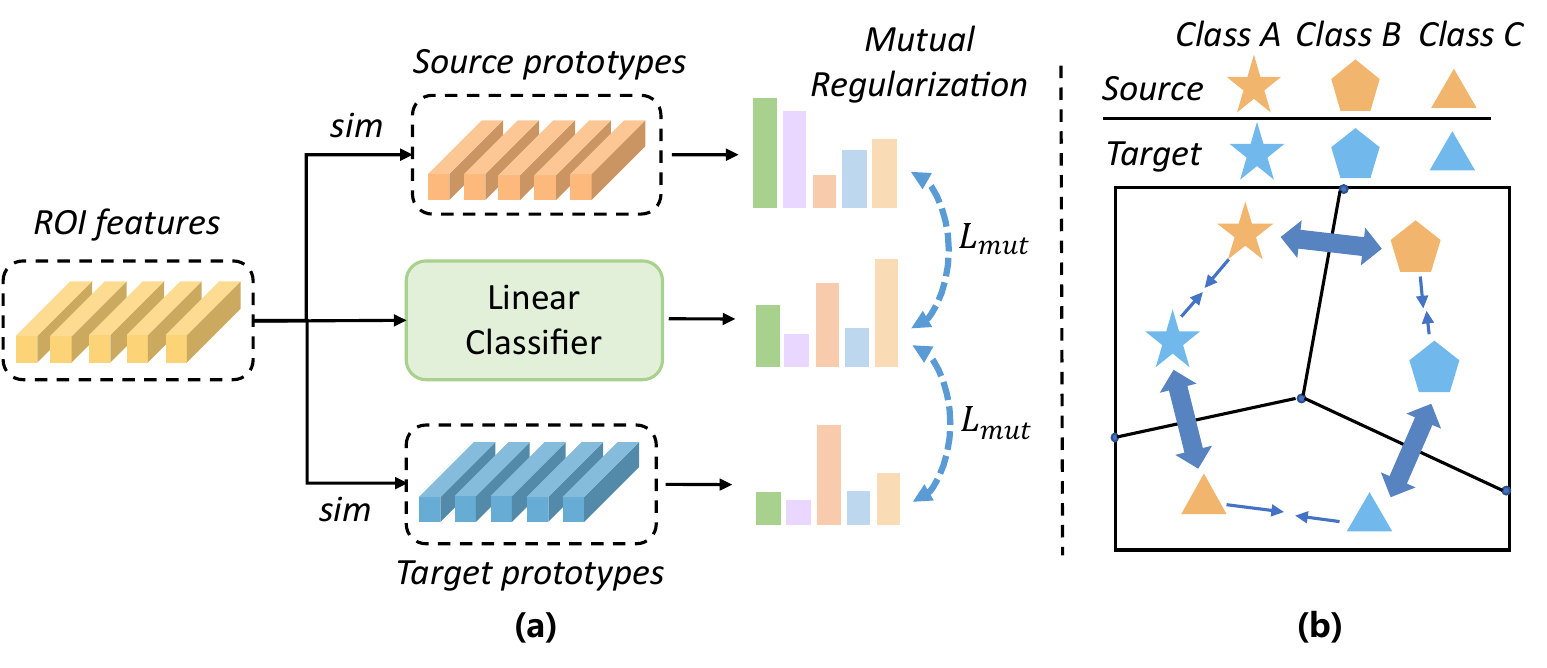}
  \caption{(a) Illustration of Prototype-based Classifier and Mutual Regularization. Source and target prototypes are utilized to classifier target ROI features, and their classifier results are conducted mutual regularization loss with linear classifier. (b) The effect of PCMR framework for source and target prototypes.}
  \label{fig:submodule}
\end{figure} 

\subsection{Prototype Cross Entropy}

\textit{Prototype} acts as the representative features for a specific class, which has been used in DAOD works ~\cite{zheng2020cross,xu2020cross}. It has been demonstrated that narrowing the distance between same categories' prototypes of two domains and increasing the distance between different classes' prototypes can improve cross-domain alignment effectiveness. However, the previous methods achieve this goal by merely calculating losses at the prototype level, lacking participation from RoI features, which cannot constrain the variance of feature distributions 
(\textit{i.e.}, even when two distributions have the same mean but different variances, they can still produce the same loss value at the prototype level).

In our method, the prototype is initialized as the average of RoI features belonging to the same class. During the training process, we directly calculate the cosine similarity between RoI features and different prototypes, as shown in Fig.~\ref{fig:submodule}(a). During the backpropagation process, loss function based on cosine similarity propagates gradients to the features, guiding them to move towards the desired distribution. Next, we will provide a theoretical analysis of the target domain features with respect to the prototype.

As discussed in Introduction, the likelihood of source RoI features is higher, whereas target RoI features show a lower likelihood. To promote performance of the detector on target domain, we choose maximizing the likelihood of target RoI features as the objective function. 

For target RoI features $\bm{x}^t$ extracted by the backbone $\bm{f}_{\theta}$, our goal is to maximize likelihood $p(\bm{x}^t;\theta)$ via optimizing the parameters $\theta$ for extracting discriminative feature representations. Assuming a set of $N$ RoI features $\{\bm{x}_i^t\}_{i=1}^N$, the optimization objective can be formulated as follows:

\begin{equation}
\begin{aligned}
\label{eq:00}
\hat{\theta}=\underset {\theta} { \operatorname {arg\,max} }\sum_{i=1}^{N}\log p(\bm{x}_i^t;\theta)
\end{aligned}
\end{equation}

The training objective is equivalent to maximizing the log-likelihood, as shown in Eq.\ref{eq:1}.
Let $\bm{\mu}_{S}$ and $\bm{\mu}_{T}$ denote the source and the target domain class prototypes corresponding to the features $\bm{x}^t$, respectively.
We first expand $p(\bm{x}^t)$ into the accumulated form of joint distribution
$p(\bm{x}^t, \bm{\mu}_{S}, \bm{\mu}_{T})$.

\begin{equation}
\begin{aligned}
\log{p(\bm{x}^t)}&=\log\sum_{\bm{\mu}_{S}}\sum_{\bm{\mu}_{T}}{p(\bm{x}^t, \bm{\mu}_{S}, \bm{\mu}_{T})}  \\
\end{aligned}  
\end{equation}

However, calculating the joint distribution is difficult. By introducing a real distribution $q(\bm{\mu}_{S}, \bm{\mu}_{T}|\bm{x}^t)$, $\log{p(\bm{x}^t)}$ can be written as:

\begin{equation}
\begin{aligned}
\log{p(\bm{x}^t)}&= \log\sum_{\bm{\mu}_{S}}\sum_{\bm{\mu}_{T}}{q(\bm{\mu}_{S}, \bm{\mu}_{T}|\bm{x}^t) \frac{p(\bm{x}^t, \bm{\mu}_{S}, \bm{\mu}_{T})}{q(\bm{\mu}_{S}, \bm{\mu}_{T}|\bm{x}^t)}} \\
&= \log \mathbb{E}_{q(\bm{\mu}_{S}, \bm{\mu}_{T}|\bm{x}^t)}[\frac{p(\bm{x}^t, \bm{\mu}_{S}, \bm{\mu}_{T})}{q(\bm{\mu}_{S}, \bm{\mu}_{T}|\bm{x}^t)}] \\
\end{aligned}  
\end{equation}

Applying Jensen's inequality,

\begin{equation}
\begin{aligned}
\log{p(\bm{x}^t)}&=\log \mathbb{E}_{q(\bm{\mu}_{S}, \bm{\mu}_{T}|\bm{x}^t)}[\frac{p(\bm{x}^t, \bm{\mu}_{S}, \bm{\mu}_{T})}{q(\bm{\mu}_{S}, \bm{\mu}_{T}|\bm{x}^t)}] \\
&\ge \mathbb{E}_{q(\bm{\mu}_{S}, \bm{\mu}_{T}|\bm{x}^t)}[\log \frac{p(\bm{x}^t, \bm{\mu}_{S}, \bm{\mu}_{T})}{q(\bm{\mu}_{S}, \bm{\mu}_{T}|\bm{x}^t)}] \\
\end{aligned}  
\end{equation}

Finally, the log-likelihood $\log{p(\bm{x}^t)}$ can be rewritten as follows:

\begin{small}
\begin{equation}
\begin{aligned}
\label{eq:1}
\log{p(\bm{x}^t)}&\!=\!\log\sum_{\bm{\mu}_{S}}\sum_{\bm{\mu}_{T}}{p(\bm{x}^t, \bm{\mu}_{S}, \bm{\mu}_{T})}  \\
&\!\ge\! \mathbb{E}_{q(\bm{\mu}_{S}\!, \bm{\mu}_{T}\!|\bm{x}^t\!)}[\log \frac{p(\bm{x}^t, \bm{\mu}_{S}, \bm{\mu}_{T})}{q(\bm{\mu}_{S}, \bm{\mu}_{T}|\bm{x}^t)}]  \\
&\!=\! \mathbb{E}_{q(\bm{\mu}_{S}\!, \bm{\mu}_{T}\!|\bm{x}^t\!)}[\log \frac{p(\bm{x}^t) p(\bm{\mu}_{S}|\bm{x}^t) p(\bm{\mu}_{T}|\bm{x}^t)}{q(\bm{\mu}_{S}|\bm{x}^t) q(\bm{\mu}_{T}|\bm{x}^t)}] \\
&\!=\! \mathbb{E}_{q(\bm{\mu}_{\hspace{-0.05em}S}\hspace{-0.05em}, \bm{\mu}_{\hspace{-0.05em}T}\!|\bm{x}^t\!)}[\log p(\!\bm{x}^t\!) \!+\! \log \frac{p(\!\bm{\mu}_{S}|\bm{x}^t\!)}{q(\!\bm{\mu}_{S}|\bm{x}^t\!)} \!+\! \log \frac{p(\!\bm{\mu}_{T}|\bm{x}^t\!)}{q(\!\bm{\mu}_{T}|\bm{x}^t\!)}]  \\
&\!=\! \mathbb{E}_{q}[\log p(\bm{x}^t)]-D_{K\!L}(q(\bm{\mu}_{S}|\bm{x}^t)\mid\mid p(\bm{\mu}_{S}|\bm{x}^t))  \\
& \quad -D_{K\!L}(q(\bm{\mu}_{T}|\bm{x}^t)\mid\mid p(\bm{\mu}_{T}|\bm{x}^t)) 
\end{aligned}  
\end{equation}
\end{small}
where $ \displaystyle p(\bm{\mu_T}=\mu_{k}|\bm{x}^t)=\frac{\exp (\cos(\mu_{k}, \bm{x}^t)/\tau)}{\sum_{i=1}^C \exp(\cos(\mu_{i},\bm{x}^t)/\tau)}$ is the posterior probability defined as the softmax of the cosine similarity between features $x^t$ and each class prototype $\mu_i$. $\mu_k$ represents the prototype of the $k$-th class, where $k$ ranges from 1 to $C$, and $C$ is the number of classes, $\tau$ is the temperature factor. Besides, $\bm{\mu}_k$ and $\bm{x}^t$ need to be normalized to the unit sphere using the $L_2$ norm. 

Here, the distribution $q(\cdot)$ represents the real distribution related to $x^t$, but the ground truth $q(\bm{\mu}_T|\bm{x^t})$ is agnostic since target data is unlabeled. Therefore, we use pseudo labels provided by the teacher model to approximate the posterior $q(\bm{\mu}_T|\bm{x^t})$. Specifically, we utilize pseudo labels $\widetilde{y^t}$ to assign category labels to target RoI features and assign a value to $q(\bm{\mu}_T|\bm{x^t})$ based on category labels, as follows:

\begin{equation}
q(\bm{\mu}_T=\mu_{k}|\bm{x^t})=
\begin{cases}
1,\quad &k=\widetilde{y^t} \\
0,\quad &k\ne\widetilde{y^t}
\end{cases} 
\end{equation}
$q(\bm{\mu}_S|\bm{x^t})$ and $p(\bm{\mu}_S|\bm{x^t})$ are calculated in the same manner.

Eq.~\ref{eq:1} provides a lower bound on the log-likelihood. To maximize this lower bound, the objective is to minimize the KL divergences with respect to the prototypes. We now define prototype cross entropy loss as:
\begin{small}
\begin{equation}
\begin{aligned}
L_{pce}&\!=\!D_{K\!L}(q(\bm{\mu}_{S}|\bm{x}^t)\!\mid\mid\! p(\bm{\mu}_{S}|\bm{x}^t))\!+\!D_{K\!L}(q(\bm{\mu}_{T}|\bm{x}^t)\!\mid\mid\! p(\bm{\mu}_{T}|\bm{x}^t)) \\
&\!=\!-\sum_k \mathbb{I}_{[k=\widetilde{y^t}]}* \left [ \log \frac{\exp (\cos(\bm{\mu}_{\bm{S}k}, \bm{x}^t)/\tau)}{\sum_{i=1}^C \exp(\cos(\mu_{\bm{S}i}, \bm{x}^t)/\tau)} \right.\\  
&\qquad\qquad\qquad\qquad \left.+\log \frac{\exp (\cos(\bm{\mu}_{\bm{T}k}, \bm{x}^t)/\tau)}{\sum_{i=1}^C \exp(\cos(\mu_{\bm{T}i}, \bm{x}^t)/\tau)}\right ]
\end{aligned}
\end{equation}
\end{small}
In traditional feature alignment, RoI features from different domains are utilized to perform global alignment without considering the semantic information. Certain class target RoI features may align with other classes, making the detector to classify incorrectly.
In our method, $L_{pce}$ encourages the target RoI features to be close to both the source and target prototypes belonging to the same class simultaneously. As shown in Fig.~\ref{fig:submodule}(b), source's and target's prototypes belonging to the same class gradually align to the same RoI features while different classes' prototypes separate from each other, resulting in an implicit alignment of the source and target features in the category-wise manner.

% Therefore, the prototype serves as a bridge connecting feature alignment and self-training, and appropriate prototypes can facilitate precise alignment and the generation of high-quality pseudo labels.

\subsection{Mutual Regularization}

The prototype cross entropy loss $L_{pce}$ encourages target domain features to become compact, which suffers from limited discriminability, resulting in suboptimal classification performance. On the other hand, the linear classifier focuses on discriminative dimensions of features, but struggles to learn compact feature representations. To enable the linear classifier to attain compactness while maintaining discriminability, we propose a mutual regularization strategy.

As illustrated in Fig.~\ref{fig:submodule}(a), this mutual regularization term imposes constraints between the linear classifier and both source and target prototype-based classifiers respectively, guiding the linear classifier to produce more compact features and promoting the prototype-based classifier leverage the linear classifier’s discriminative power. We employ Jensen-Shannon (JS) divergence as the loss function. The mutual regularization loss is formulated as follows:
\begin{equation}
\begin{aligned}
  L_{mut}=D_{\rm{JS}}(p_{\rm{lin}}(\bm{y}|\bm{x}^t)||p_{\rm{pro}}^{s}(\bm{\mu}|\bm{x}^t)) \\ +
  D_{\rm{JS}}(p_{\rm{lin}}(\bm{y}|\bm{x}^t)||p_{\rm{pro}}^{t}(\bm{\mu}|\bm{x}^t))
\end{aligned}
\end{equation}
where $p_{\rm{pro}}(\bm{\mu}|\bm{x}^t)$ and $p_{\rm{lin}}(\bm{y}|\bm{x}^t)$ represent the prediction of the prototype and linear classifier, respectively.

The overall loss of the detector is as follows:
\begin{equation}
  L=L_{AT}+\lambda_{1}\cdot L_{pce}+\lambda_{2}\cdot L_{mut}
\end{equation}
where $\lambda_{1}$ and $\lambda_{2}$ are hyper-parameters, which trade off $L_{pce}$ and $L_{mut}$.

By integrating this regularization loss with other objectives, the detector can achieve a reasonable trade-off between compactness and discriminability in feature space.

\subsection{Prototype Update Strategy}
\label{subsec:prototype}
In this section, we will introduce the initialization and updating methods for prototypes.

We utilize the detector trained on the source training data to perform inference on the target domain. We average the RoI features with classification scores above the confidence threshold $T$ as the prototype. The initial prototype for the $k$-th class can be calculated as:
\begin{equation}
\label{eq:init}
  \bm{\mu}_k^0 = \frac{\sum_{i}^{N} \mathbb{I}_{\left [p(y^k|x_i)\ge T\right ]}\cdot x}{\sum_{i}^{N} \mathbb{I}_{\left [p(y^k|x_i)\ge T\right ]}}
\end{equation}
We use the same process to obtain the source prototype, except for inference on the source domain.

In our approach, source and target prototypes are dynamically updated based on the RoI features corresponding to source ground-truth labels and target pseudo labels respectively during the training process.
First, for a mini-batch, we average RoI features as the mini-batch prototypes [$\widetilde{\bm{\mu}_1}, \dots, \widetilde{\bm{\mu}_k}$] in the category-wise manner. Considering RoI features may contain noise in the early training stage, to obtain a more robust and precise prototype, we calculate the cosine similarity between prototype at $(t-1)_{th}$ iteration $\bm{\mu}_k^{t-1}$ and $\widetilde{\bm{\mu}_k}$, and map it to [0, 1] to serve as a weighted coefficient $\alpha$, as follows:

\begin{equation}
  \alpha = \frac{{\rm sim}(\bm{\mu}_k^{t-1}, \widetilde{\bm{\mu}_k})+1}{2}
\end{equation}

Subsequently, we perform a weighted average using the $\widetilde{\bm{\mu}_k}$ and $\bm{\mu}_k^{t-1}$:

\begin{equation}
  \bm{\mu}_k^t = (1-\alpha)\cdot \bm{\mu}_k^{t-1}+\alpha\cdot \widetilde{\bm{\mu}_k}
\end{equation}

Finally, we normalize the prototypes $\bm{\mu}_k^t$. The overall update process can be referred to Alg.~\ref{alg:prototype}.

\begin{figure}[h]
  %\removelatexerror
  \begin{algorithm}[H]
      \caption{Prototype update algorithm}
      \label{alg:prototype}
      \begin{algorithmic}[1]
          \REQUIRE Input image $I$ with GT or pseudo label set $P$
          \ENSURE prototype $\mu_k$ for $k$-th class \\
          % \STATE Initialize prototype [${\mu_1}, \dots, {\mu_K}$]
          \STATE Feature map \textit{E}$\gets$ Backbone($I$)
          % \STATE Proposals \textit{P}$\gets$ GT or pseudo labels
          \STATE RoI features $f$ $\gets$ RoI(\textit{E, P}).
          \STATE [$\widetilde{\mu_1}, \!\dots\!, \widetilde{\mu_K}$] $\gets$ Calculate mini-batch prototype using $f$
          \FOR {$i=1$ to $K$}
            \STATE $\alpha\gets$ Calculate Cosine Similarity $\widetilde{\mu_i}$ and ${\mu_i}$
            \STATE $\alpha\gets(\alpha+1)/2$
            \STATE $\mu_i\gets (1-\alpha)*\mu_i+\alpha*\widetilde{\mu_i}$
            \STATE $\displaystyle \mu_i\gets \frac{\mu_i} {\| \mu_i\|_2}$
          \ENDFOR
          \STATE {\textbf{return} [${\mu_1}, \dots, {\mu_K}$]}
      \end{algorithmic}
  \end{algorithm}
\end{figure}

% *--***************************************
\begin{table*}[!htbp]
    \caption{Quantitative results from Cityscapes$\rightarrow$Foggy Cityscapes (`0.02' split) based on different base detectors with various backbones.$^\dagger$ Results reproduced using the released code by CMT~\cite{cao2023contrastive}.}.
    \small
        \centering
         % \resizebox{1.0\textwidth}{!}{%
        \begin{tabular}{>{\raggedright}p{3.0cm}|p{1.3cm}<{\centering}|p{1cm}<{\centering}p{1cm}<{\centering}p{1cm}<{\centering}p{1cm}<{\centering}p{1cm}<{\centering}p{1cm}<{\centering}p{1cm}<{\centering}p{1cm}<{\centering}p{1cm}<{\centering}}
        \hline
        Method & Backbone & person & rider & car & truck & bus & train & mcycle & bcycle & mAP \\ \hline \hline
        % SWDA \cite{saito2019strong}  & \multirow{14}{*}{VGG16} & 36.2 & 35.3 & 43.5 & 30.0 & 29.9 & 42.3 & 32.6 & 24.5 & 34.3 \\
        % TIA \cite{zhao2022task} &  & 34.8 & 46.3 & 49.7 & 31.1 & 52.1 & 48.6 & 37.7 & 38.1 & 42.3 \\
        TDD \cite{he2022cross} & \multirow{11}{*}{VGG16} & 39.6 & 47.5 & 55.7 & 33.8 & 47.6 & 42.1 & 37.0 & 41.4 & 43.1 \\
        % MGA \cite{zhou2022multi} &  & 45.7 & 47.5 & 60.6 & 31.0 & 52.9 & 44.5 & 29.0 & 38.0 & 43.6 \\
        PT \cite{chen2022learning} &  & 40.2 & 48.8 & 59.7 & 30.7 & 51.8 & 30.6 & 35.4 & 44.5 & 42.7 \\
        SCAN \cite{li2022scan} &  & 41.7 & 43.9 & 57.3 & 28.7 & 48.6 & 48.7 & 31.0 & 37.3 & 42.1 \\
        SIGMA++ \cite{li2023sigma++} &  & 46.4 & 45.1 & 61.0 & 32.1 & 52.2 & 44.6 & 34.8 & 39.9 & 44.5 \\
        HT \cite{deng2023harmonious} &  & 52.1 & 55.8 & 67.5 & 32.7 & 55.9 & 49.1 & 40.1 & 50.3 & 50.4 \\
        % OADA \cite{yoo2022unsupervised} &  & 47.8 & 46.5 & 62.9 & 32.1 & 48.5 & 50.9 & 34.3 & 39.8 & 45.4 \\
        SIGMA \cite{li2022sigma} &  & 43.9 & 52.7 & 56.8 & 26.2 & 46.2 & 12.4 & 34.8 & 43.0 & 43.5 \\
        MGA \cite{zhang2024robust} &  & 47.0 & 54.6 & 64.8 & 28.5 & 52.1 & 41.5 & 40.9 & 49.5 & 47.4 \\

        AT \cite{li2022cross} &  &  45.3 & 55.7 & 63.6 & 36.8 & 64.9 & 34.9 & 42.1 & 51.3 & 49.3  \\ 
        \cellcolor{lightgray!45}AT \cite{li2022cross} + PACF  & &  \cellcolor{lightgray!45}45.4 & \cellcolor{lightgray!45}57.4 & \cellcolor{lightgray!45}63.9 & \cellcolor{lightgray!45}38.0 &  \cellcolor{lightgray!45}61.5 &  \cellcolor{lightgray!45}51.3 & \cellcolor{lightgray!45}42.9 & \cellcolor{lightgray!45}52.5 & \cellcolor{lightgray!45}51.6  \\  
        CMT \cite{cao2023contrastive} & & 45.9 & 55.7 & 63.7 & 39.6 &66.0 & 38.8 & 41.4 & 51.2 & 50.3  \\ 
        % CMT+DSD-DA \cite{fengdsd} & & 49.0 & 59.6 & 65.3 & 35.7 & 61.0 & 46.5 & 43.9 & 57.3 & 52.3 \\
         \cellcolor{lightgray!45}CMT \cite{cao2023contrastive} \!+\! PACF  & &   \cellcolor{lightgray!45}45.6 &  \cellcolor{lightgray!45}57.1 &  \cellcolor{lightgray!45}63.2 &  \cellcolor{lightgray!45}38.7 &   \cellcolor{lightgray!45}62.5 &   \cellcolor{lightgray!45}53.6 &  \cellcolor{lightgray!45}44.2 &  \cellcolor{lightgray!45}53.6 &  \cellcolor{lightgray!45}\textbf{52.3}  \\  
        \hline \hline
        % \cline{1-1} \cline{3-11}
        GPA \cite{xu2020cross} & \multirow{8}{*}{ResNet50} & 32.9 & 46.7 & 54.1 & 24.7 & 45.7 & 41.1 & 32.4 & 38.7 & 39.5 \\
        % CRDA \cite{xu2020exploring} &  & 39.9 & 38.1 &57.3 & 28.7 & 50.7 & 37.2 & 30.2 & 34.2 & 39.5 \\
        % DIDN \cite{lin2021domain} &  & 38.3 & 44.4 & 51.8 & 28.7 & 53.3 & 34.7 & 32.4 & 40.4 & 40.5 \\
        % DSS \cite{wang2021domain} &  & 42.9 & 51.2 & 53.6 & 33.6 & 49.2 & 18.9 & 36.2 & 41.8 & 40.9 \\
        MIC \cite{hoyer2023mic}   &  & 50.9 & 55.3 & 67.0 & 33.9 & 52.4 & 33.7 & 40.6 & 47.5 & 47.6 \\
        DSD-DA \cite{fengdsd}     &  & 43.7 & 49.1 & 60.7 & 30.8 & 55.7 & 43.4 & 33.7 & 44.6 & 45.2 \\
        MTM \cite{weng2024mean}   &  & 51.0 & 53.4 & 67.2 & 37.2 & 54.4 & 41.6 & 38.4 & 47.7 & 48.9 \\
        
        AT $^\dagger$\cite{li2022cross} &  & 48.3 & 56.4 & 66.0 & 37.6 & 61.0 & 38.2 & 42.1 & 53.0 & 50.3  \\ 
        \cellcolor{lightgray!45}AT \cite{li2022cross} + PACF  & &  \cellcolor{lightgray!45}47.4  & \cellcolor{lightgray!45} 57.3 & \cellcolor{lightgray!45} 65.7 & \cellcolor{lightgray!45} 38.8 &  \cellcolor{lightgray!45} 62.3 &  \cellcolor{lightgray!45} 44.7 & \cellcolor{lightgray!45} 43.0 & \cellcolor{lightgray!45} 51.4 & \cellcolor{lightgray!45}51.3  \\  
        CMT $^\dagger$\cite{cao2023contrastive} &  & 47.9 & 56.3 & 65.1 & 36.2 & 62.4 & 43.4 & 41.3 & 52.2 & 50.6 \\ 
        
        \cellcolor{lightgray!45}CMT \cite{cao2023contrastive} \!+\! PACF  & &  \cellcolor{lightgray!45}47.1  & \cellcolor{lightgray!45}57.3 & \cellcolor{lightgray!45}65.1 & \cellcolor{lightgray!45}38.2 &  \cellcolor{lightgray!45}61.7 &  \cellcolor{lightgray!45}47.5 & \cellcolor{lightgray!45}42.4 & \cellcolor{lightgray!45} 50.8 & \cellcolor{lightgray!45}\textbf{51.4}  \\  
        \hline \hline
        
        % CADA \cite{hsu2020every} & \multirow{6}{*}{ResNet101} & 41.5 & 43.6 & 57.1 & 29.4 & 44.9 & 39.7 & 29.0 & 36.1 & 40.2 \\
        % D-adapt \cite{jiang2021decoupled} & \multirow{6}{*}{ResNet101} & 42.8 & 48.4 & 56.8 & 31.5 & 42.8 & 37.4 & 35.2 & 42.4 & 42.2 \\
        DSD-DA \cite{fengdsd}      & \multirow{6}{*}{ResNet101} & 43.9 & 50.7 & 61.6 & 31.8 & 52.2 & 47.1 & 32.1 & 46.1 & 45.7 \\
        MGA \cite{zhang2024robust} &  & 47.2 & 48.1 & 63.7 & 37.5 & 54.6 & 50.8 & 28.8 & 44.2 & 46.9 \\
    
        % \cline{1-1} \cline{3-11}
        AT $^\dagger$\cite{li2022cross} &  & 48.1 & 54.8 & 65.5 & 40.6 & 61.0 & 41.6 & 38.3 & 50.4 & 50.0  \\ 
        \cellcolor{lightgray!45}AT \cite{li2022cross} + PACF  & &  \cellcolor{lightgray!45} 47.6 & \cellcolor{lightgray!45} 54.6 & \cellcolor{lightgray!45} 65.1 & \cellcolor{lightgray!45} 42.3 &  \cellcolor{lightgray!45} 63.4 &  \cellcolor{lightgray!45} 42.5 & \cellcolor{lightgray!45} 39.3 & \cellcolor{lightgray!45} 52.4 & \cellcolor{lightgray!45}50.9  \\  
        
        CMT $^\dagger$\cite{cao2023contrastive} & & 48.4 & 55.8 & 65.9 & 41.5 & 60.3 & 37.0 & 36.0 & 51.1 & 49.5 \\ 
        \cellcolor{lightgray!45}CMT \cite{cao2023contrastive} \!+\! PACF  & &  \cellcolor{lightgray!45} 47.5 & \cellcolor{lightgray!45} 56.4 & \cellcolor{lightgray!45} 65.2 & \cellcolor{lightgray!45} 38.8 &  \cellcolor{lightgray!45} 59.5 &  \cellcolor{lightgray!45} 50.5 & \cellcolor{lightgray!45} 41.4 & \cellcolor{lightgray!45} 50.9 & \cellcolor{lightgray!45}\textbf{51.3}  \\
       \hline
        \end{tabular}
        \label{table:foggy}
        
\end{table*}

\begin{table}[!htbp]
    \caption{Quantitative results from Sim10k (S)$\rightarrow$Cityscapes (C) and Cityscapes (C)$\leftrightarrows$KITTI (K) based on Faster RCNN with VGG16 backbone.}
    \small
        \centering
         % \resizebox{0.48\textwidth}{!}{%
        % \begin{tabular}{>{\raggedright}p{2.0cm}|p{3.0cm}<{\centering}|p{3.0cm}<{\centering}|p{3.0cm}<{\centering}}
        % \begin{tabular}{c|c|c|c}
        \begin{tabular}{p{3cm}<{\centering}|p{1.2cm}<{\centering}|p{1.2cm}<{\centering}|p{1.2cm}<{\centering}}
        \hline
        Method & S$\rightarrow$C & K$\rightarrow$C  & C$\rightarrow$K \\ \hline \hline
        MEGA \cite{vs2021mega} & 44.8 & 43.0 & 75.5   \\
        RPNPA \cite{zhang2021rpn} & 45.7 & - & 75.1   \\
        TIA \cite{zhao2022task} & - & 44.0 & 75.9   \\
        % SCDA \cite{zhu2019adapting} & 43.0 & 42.5  & -   \\
        % HTCN \cite{chen2020harmonizing} & 42.5 & 42.1 & -   \\
        UMT \cite{deng2021unbiased} & 43.1 & - & -  \\
        SED \cite{li2021free} & 42.5 & 43.7 & -   \\
        TDD \cite{he2022cross} & 53.4 & 47.4 & -  \\
        MGA \cite{zhou2022multi} & 54.6 & 48.5 & -  \\

        AT \cite{li2022cross} & 54.6 & 46.1 & 74.5  \\
         \cellcolor{lightgray!45} AT \cite{li2022cross} + PACF &  \cellcolor{lightgray!45}56.7 &  \cellcolor{lightgray!45}47.5 &  \cellcolor{lightgray!45}75.7   \\
        CMT \cite{cao2023contrastive} & 56.2 & 47.4 & 75.5   \\
         \cellcolor{lightgray!45} CMT \cite{cao2023contrastive} + PACF &  \cellcolor{lightgray!45}\textbf{56.7} &  \cellcolor{lightgray!45}\textbf{47.6} & \cellcolor{lightgray!45}\textbf{76.4}   \\
    \hline
        \end{tabular}
        % }
        \label{table:combine}
        
\end{table}

\section{Experiments}
\label{sec:blind}

In this section, we first introduce our experimental settings. Then we show the comparisons with state-of-the-art DAOD detectors. Finally,
comprehensive experiments with detailed ablation studies are conducted on PACF to show the effectiveness of each component, \textit{i.e.}, prototype cross entropy and mutual regularization.

\subsection{Domain Adaptation Settings}
 
\textbf{\textit{Cityscapes$\rightarrow$FoggyCityscapes}}. \textbf{\textit{Cityscapes}} dataset comprises street scenes under clear weather conditions using an onboard camera. In the DAOD task, this dataset is divided into a training set (2975 images) and a validation set (500 images), with bounding boxes annotated for eight categories. Building upon \textbf{\textit{Cityscapes}}, \cite{sakaridis2018semantic} introduced \textbf{\textit{FoggyCityscapes}} by incorporating heavy foggy conditions. Following prior work, we focus on the most challenging scenario (0.02 fog level) as the target domain to investigate the domain gap induced by weather conditions in this adaptation from clear to foggy environments.

\begin{table}[!t]
    \caption{Ablation studies of components in our PACF framework.}
    \small
    \centering
    % \vspace{-1.5pt}
    % \resizebox{0.48\textwidth}{!}{%
    \renewcommand{\arraystretch}{1.2}
    \begin{tabular} {c|p{0.8cm}<{\centering}|p{0.6cm}<{\centering} p{0.6cm}<{\centering} p{0.6cm}<{\centering}|c}
      \hline
       \multirow{2}{*}{Method} &  \multirow{2}{*}{$L_{pce}$} & \multicolumn{3}{c|}{Prototype Regularization}  & \multirow{2}{*}{AP} \\
      &  & $L_2$ & $\rm{KL}$ & $\rm{JSD}$  & \\
      \hline
      Baseline & - & - & -& -& 49.3 \\
      \hline
      1 & & \checkmark & & & 51.1\textcolor{red}{(+1.8)} \\
      2 & & & \checkmark & & 50.9\textcolor{red}{(+1.6)}\\
      3 & & &  & \checkmark &  50.6\textcolor{red}{(+1.3)}\\
      4 & \checkmark & & & & 50.9\textcolor{red}{(+1.6)} \\
      5 & \checkmark & \checkmark & & & 50.8\textcolor{red}{(+1.5)} \\
      6 & \checkmark & & \checkmark & &  51.2\textcolor{red}{(+1.9)}\\
      7 & \checkmark & &  & \checkmark & 51.6\textcolor{red}{(+2.3)} \\
      \hline
    \end{tabular}
    \label{tab:loss}
    % }
  \end{table}

\textbf{\textit{Sim10k$\rightarrow$Cityscapes}}. \textbf{\textit{Sim10k}}~\cite{johnson2016driving} dataset is a synthetic dataset, presenting an inherent domain gap when compared to the real-world \textbf{\textit{Cityscapes}} dataset. \textbf{\textit{Sim10k}} comprises 10,000 images with labeled bounding boxes in the car class. In line with prior research, we undertake adaptation in this synthetic-to-real scenario, evaluating the performance on the car class.

\textbf{\textit{KITTI$\leftrightarrows$Cityscapes}}. 
\textbf{\textit{KITTI}}~\cite{geiger2012we} dataset consists of real-world traffic scenes captured by vehicle-mounted cameras, presenting an inherent domain gap when compared with \textbf{\textit{Cityscapes}}, which is captured from onboard cameras. \textbf{\textit{KITTI}} includes annotated cars in 7,481 images, highlighting a cross-viewpoint domain disparity. We follow the literature to explore dual-directional adaptations between these two datasets to fully evaluate our algorithm.

\subsection{Implementation Details}

For a fair comparison with previous methods, we use the standard Faster R-CNN object detector~\cite{ren2015faster} with the VGG16~\cite{simonyan2014very} and ResNet backbone as the detection model. As for hyperparameters in all experiments, we set $\lambda_{1}=1$, $\lambda_{2}=1$, and the temperature $\tau=0.05$, $T=0.8$.
Other hyper-parameters are the same as in the original implementation of AT~\cite{li2022cross}. Additionally, for single class adaptation setting, we calculate the prototype-based class probabilities using sigmoid activation function, instead of softmax.
Except for AT, we also adopt CMT~\cite{cao2023contrastive} as our baseline. In the inference stage, we only adopt the linear classifier to obtain prediction results.
Our implementation is based on Detectron2~\cite{wu2019detectron2} and the publicly available code by AT and CMT. 
Each experiment is conducted on 4 NVIDIA 3090 GPUs.

\subsection{Comparison with State-of-the-Arts}

We compare our PACF with several state-of-the-art methods, validating the effectiveness of our approach under three different domain adaptation scenarios.

\textbf{Weather Conditions}.
Table~\ref{table:foggy} shows the results from \textit{Cityscapes} to \textit{FoggyCityscapes}. 
Our method achieves consistent improvement using different backbones with two baselines, \textit{e.g.}, with VGG16 as backbone, our method outperforms state-of-the-art method (\textit{i.e.}, CMT) by 2.0\%. In particular, a significant improvement is observed in the `train' category. Due to the similarity between `train' and `truck' in the dataset, they are prone to misclassification. The prototype cross entropy contributes to extracting more compact feature representation, making these two class features separate from each other, leading to a substantial improvement in the classification of similar categories.

\textbf{Synthetic to Real}. 
This adaptation process enhances training data by matching features from synthetic images to the distribution of real images, boosting the model's generalization performance. Table~\ref{table:combine} provides the results from \textit{Sim10k$\rightarrow$Cityscapes}. PACF achieves 56.7\% AP, exceeding the baseline by 2.1\% and demonstrating robust generalization from synthetic to real images.

\textbf{Across Cameras}. Variations in camera angles and imaging devices (\textit{e.g.}, different intrinsic parameters) during capture contribute to domain shifts. We conduct experiments in the \textit{Cityscapes$\rightarrow$KITTI} setting. As shown in Table~\ref{table:combine}, our method outperforms the baseline by 1.2\% AP. In the \textit{KITTI$\rightarrow$Cityscapes} setting, our result is 47.5\% AP, surpassing the baseline by 1.4\%.

\subsection{Ablation Study}

In this section, we conduct ablation experiments in the adaptation setting from \textit{Cityscapes} to \textit{FoggyCityscapes}. This setting involves multiple categories' adaptation, effectively showing the effectiveness and generality of our approach.

\begin{figure*}[!htbp]
  \centering
    \includegraphics[width=0.9\textwidth]{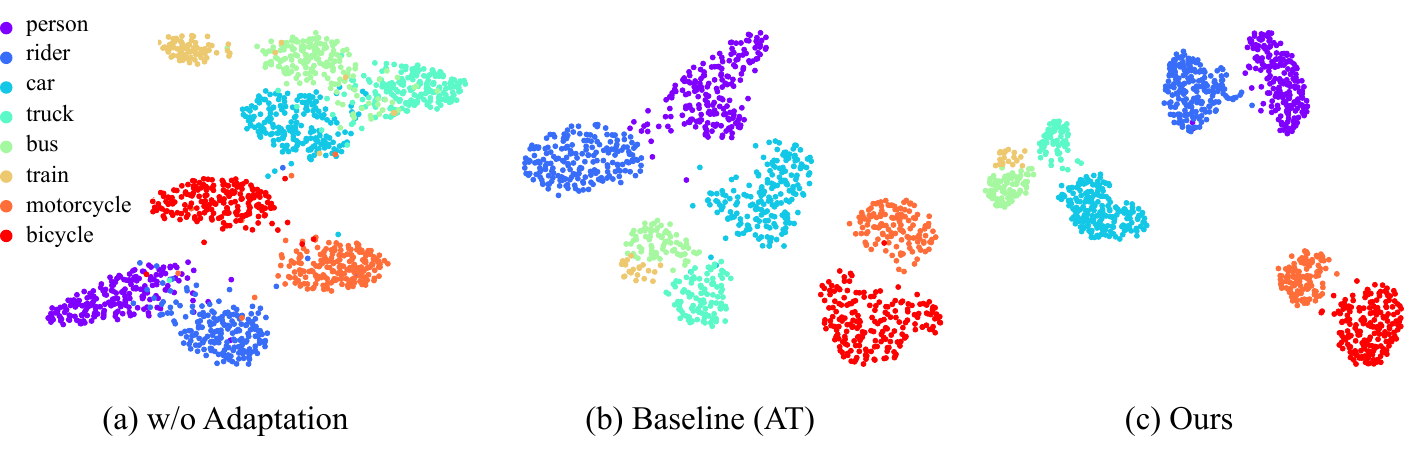}
    \caption{t-SNE visualization of target RoI features in the \textit{Cityscapes$\rightarrow$FoggyCityscapes} setting.
            }
  \label{fig:visualization}
\end{figure*}
\begin{table*}[!htbp]
    \caption{Quantitative results of the variance of features across different classes in the source and target domains. `AT + Inst' represents the method of adding instance-level feature alignment on AT.}
    \small
    \centering
    % \resizebox{0.7\textwidth}{!}{
    \setlength{\tabcolsep}{9.5pt}
    \begin{tabular}{ccc|cc|cc|cc|cc}
    \hline
    \multirow{2}{*}{Category} & \multicolumn{2}{c}{w/o Adaptation} & \multicolumn{2}{c}{AT} & \multicolumn{2}{c}{AT+Inst} & \multicolumn{2}{c}{CMT} & \multicolumn{2}{c}{Ours} \\
     & \multicolumn{1}{c}{Source} & \multicolumn{1}{c}{Target} & \multicolumn{1}{c}{Source} & \multicolumn{1}{c}{Target} & \multicolumn{1}{c}{Source} & \multicolumn{1}{c}{Target} & \multicolumn{1}{c}{Source} & \multicolumn{1}{c}{Target} & \multicolumn{1}{c}{Source} & \multicolumn{1}{c}{Target} \\
    \hline
    \multicolumn{1}{c|}{person} & 5.315 & 5.216 & 4.420 & 4.258 & 3.668 & 3.826 & 4.778 & 4.858 & 2.739 & 2.695 \\
    \multicolumn{1}{c|}{rider} & 13.969 & 13.863 & 10.220 & 10.872 & 9.940 & 9.806 & 11.163 & 11.820 & 7.238 & 7.354 \\
    \multicolumn{1}{c|}{car} & 6.606 & 6.820 & 5.752 & 5.633 & 5.129 & 5.265 & 6.334 & 6.432 & 2.989 & 3.021 \\
    \multicolumn{1}{c|}{truck} & 19.872 & 18.705 & 9.443 & 9.925  & 14.342 & 13.203 & 19.265 & 16.889 & 11.185 & 9.919 \\
    \multicolumn{1}{c|}{bus} & 16.432 & 17.202 & 9.642 & 9.424 & 8.526 & 10.549 & 13.698 & 12.746 & 6.468 & 8.035 \\
    \multicolumn{1}{c|}{train} & 21.803 & 47.971 & 10.471 & 9.041 & 13.138 & 12.037  & 16.614 & 19.953 & 9.258 & 12.015 \\
    \multicolumn{1}{c|}{motorcycle}    & 13.890 & 13.955 & 9.051 & 9.384 & 10.614 & 9.962 & 14.154 & 14.253 & 5.631 & 6.729 \\
    \multicolumn{1}{c|}{bicycle} & 12.956 & 11.520 & 9.421 & 8.485 & 8.231 & 8.177 & 9.958 & 10.060 & 6.176 & 6.077 \\
    \hline
    \multicolumn{1}{c|}{avg.} & 13.856 & 16.906 & 8.552 & 8.378 & 9.198 & 9.103 & 12.008 & 12.126 & \textbf{6.460} & \textbf{6.980} \\
    \hline
    \end{tabular}
    % }
    \label{table:var}
    \end{table*}
\begin{table}[!tbp]
    \caption{Quantitative results of the mean shifts between the source and target domains. `AT + Inst' represents the method of adding instance-level feature alignment on AT.}
    \small
    % \resizebox{0.48\textwidth}{!}{
    \centering
    % \begin{tabular} {c|p{0.8cm}<{\centering}|p{0.6cm}<{\centering} p{0.6cm}<{\centering} p{0.6cm}<{\centering}|c}
    \begin{tabular}{c|c|c|c|c|c}
    % \begin{tabular} {c|p{1.8cm}<{\centering}|p{1.0cm}<{\centering}|p{1.0cm}<{\centering}|p{1.0cm}<{\centering}|p{1.0cm}<{\centering}}
    \hline
    \multicolumn{1}{l}{Category} & \multicolumn{1}{c}{w/o Adaptation} & \multicolumn{1}{c}{AT} & \multicolumn{1}{c}{AT+Inst} & \multicolumn{1}{c}{CMT} & \multicolumn{1}{c}{Ours} \\
    \hline
    person & 0.632 & 0.495 & 0.152 & 0.189 & 0.142 \\
    rider & 0.874 & 0.857 & 0.393& 0.405 & 0.311 \\
    car & 0.653 & 0.514 & 0.136 & 0.169 & 0.083 \\
    truck & 2.831 & 1.562 & 0.817 & 1.363 & 0.890 \\
    bus & 3.629 & 1.887 & 0.768 & 1.351 & 0.662 \\
    train & 6.051 & 2.200 & 3.330 & 2.548 & 1.786 \\
    motor & 1.542 & 0.690 & 0.754 & 0.812 & 0.621 \\
    bicycle & 0.643 & 0.530 & 0.197 & 0.244 & 0.188 \\
    \hline
    avg. & 2.107 & 1.092 & 0.818 & 0.885 & \textbf{0.585} \\
    \hline
    \end{tabular}
    % }
    \label{table:mean}
    \end{table}

\textbf{Effectiveness of each component}. First, we demonstrate the effectiveness of each component. The results are shown in Table~\ref{tab:loss}. 
Rows 1-3 in Table~\ref{tab:loss} indicate that during training, we simultaneously employ two types of classifiers and use $L_{mut}$ to enhance correlation between these classifiers' predictions, but we do not use $L_{pce}$ to supervise the prototype-based classification results. We can observe that directly applying the mutual regularization loss on the baseline can improve performance consistently. Among the regularization losses, the $L_2$ loss yields the highest improvement, surpassing baseline by 1.8\% AP. 
Furthermore, when we solely use $L_{pce}$, the model exhibits a relative improvement of 1.6\% over baseline, highlighting the positive impact of prototypes on extracting compact features.
Finally, with all these components, we achieve the highest detection performance of 51.6\% AP. 

\begin{table}[t]
    \caption{True Positive (TP) ratio of pseudo labels for each category between the baseline and our method.}
    \resizebox{0.48\textwidth}{!}{
    \begin{tabular}{c|c|c|c|c|c|c|c|c|c}
    \hline
    \multicolumn{1}{c}{Method} & \multicolumn{1}{l}{prsn} & \multicolumn{1}{l}{rider} & \multicolumn{1}{l}{car} & \multicolumn{1}{l}{truc} & \multicolumn{1}{l}{bus} & \multicolumn{1}{l}{tran} & \multicolumn{1}{l}{moto}& \multicolumn{1}{l}{bick} & \multicolumn{1}{l}{avg.}\\
    \hline
    AT (Baseline) & 0.58          & \textbf{0.82} & 0.75          & \textbf{0.62} & 0.79          & 0.64                   & 0.65        & 0.65          & 0.688 \\
    Ours          & \textbf{0.65} & 0.81          & \textbf{0.84} & 0.60          & \textbf{0.80} & \textbf{0.67} & \textbf{0.66} & \textbf{0.71} & \textbf{0.718} \\
    \hline
    \end{tabular}
    }
    \label{table:acc}
    % \vspace{-0.4cm}
    \end{table}

\begin{figure}[t]
  \centering
  \includegraphics[width=0.50\textwidth]{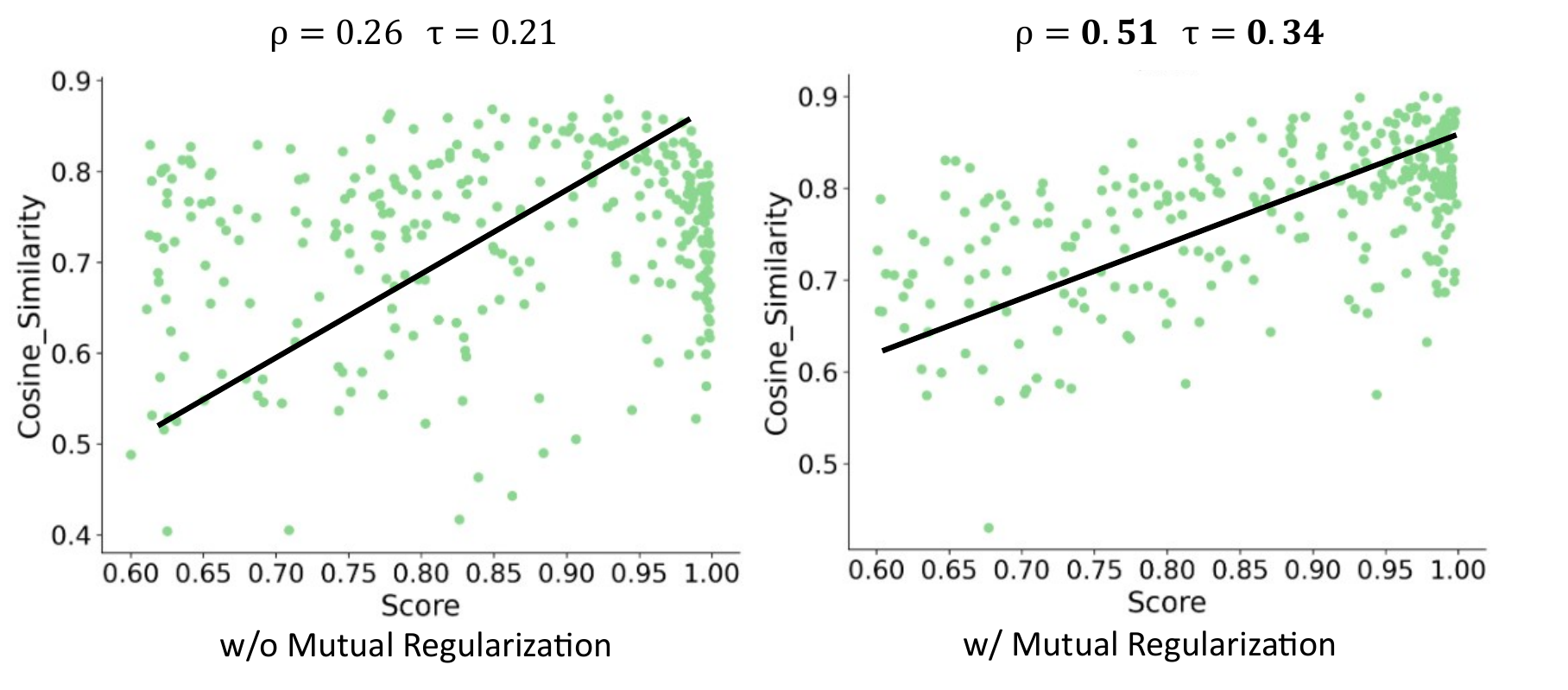}
  \caption{The correlation between linear classification scores and prototype-based cosine similarity. Left and right represent without and with mutual regularization.}
  \label{fig:scatter}
  \vspace{-0.5cm}
\end{figure}

\textbf{Impact on feature distribution}. 
The core idea of proposing $L_{pce}$ is to utilize the prototype for alleviating mean shift and large intra-class variation problem. Therefore, we analyze the influence of $L_{pce}$ on feature distribution. Fig.~\ref{fig:visualization} presents the feature distribution of 8 classes in FoggyCityscapes. 
We can find that the model without adaptation is prone to class confusion and shows high variance. The baseline reduces misclassification, yet the variance remains relatively high. Our method tends to produce more compact feature distributions for every class, which will ease knowledge transfer from source to target.
For quantitative comparison, we present the class-wise variances in source and target domains under different methods in Table~\ref{table:var}. It can be observed that our method significantly reduces variances in each class. 
Additionally, Table~\ref{table:mean} demonstrates that our method exhibits smaller mean shift, leading to a relative reduction of 46\% compared to the baseline.

Furthermore, we investigate the influence of instance-level feature alignment~\cite{chen2018domain} via adversarial learning on class-conditional distribution. 
It is worth noting that in Table~\ref{table:var}, adversarial learning tends to make the intra-class variance larger compared with baseline. The reason can be that this instance-level alignment is performed in the category-agnostic manner, which leads to certain class's distribution is misaligned to other multi-classes' distribution.

\begin{figure*}[!htbp]
	\centering
        \includegraphics[width=1.00\textwidth]{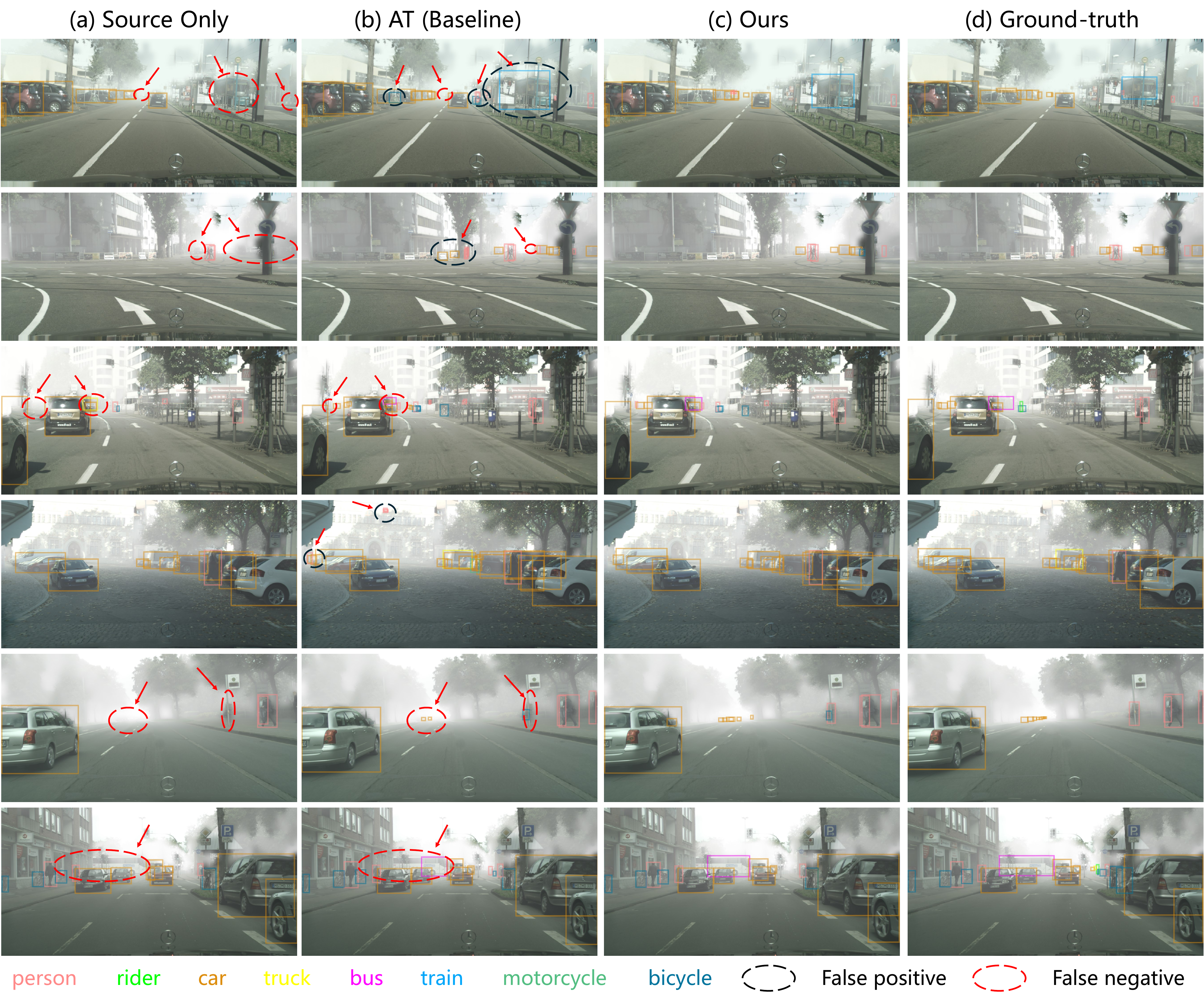}
	\caption{Qualitative results on the \textbf{\textit{Cityscapes-to-FoggyCityscapes}} adaptation scenario of (a) the \textit{Source Only} model, (b) \textit{Baseline} \cite{li2022cross}, (c) \textit{Ours}, and (d) \textit{Ground-truth}. (Zooming in for best view.)}
 \label{fig:results}
\end{figure*}
\begin{figure}[!htbp]
   \centering
   \includegraphics[width=1.00\linewidth]{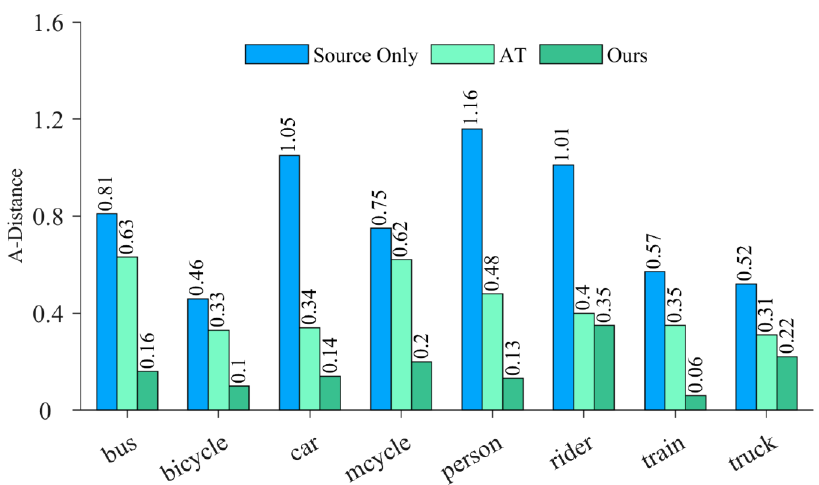}
   \caption{Feature distribution discrepancy of foregrounds.}
   \label{fig:distance} 
\end{figure}

\textbf{Effectiveness of mutual regularization}. 
In this paper, we introduce Spearman Rank Correlation Coefficient $\rho$ and Kendall Rank Correlation Coefficient $\tau$ as the measures of rank consistency: the similarity of the orderings of the data when ranked by each of the quantities. 
Fig.~\ref{fig:scatter} shows that $L_{mut}$ can promote the correlation between two types of classifiers. Under this constraint, the linear classifier and prototype-based classifier tend to produce consistent predictions, thereby balancing feature compactness and discriminability.

In addition, we compare the pseudo label quality between baseline and our method. Table~\ref{table:acc} shows our method has higher true positive ratio. 
During the training stage, there exists that prototype-based classifiers produce more accurate and confident prediction than linear classifier. As for these predictions, through mutual regularization strategy, linear classifier can improve its classification scores above pseudo-labeled threshold to align with prototype-based classifiers. This leads to these high-quality predictions are filtered as pseudo labels, which effectively expand pseudo labels.  

\subsection{Further Analysis}

\textbf{Distribution discrepancy of foregrounds.}
Theoretical results in~\cite{ben2006analysis} suggest that the $\mathcal{A}$-distance serves as an effective metric for quantifying domain discrepancy. In practical applications, we compute the Proxy $\mathcal{A}$-distance as an approximation, defined as $d_{\mathcal{A}} = 2(1 - \epsilon)$, where $\epsilon$ represents the generalization error of a binary classifier used to discriminate the origin domain of RoI features.
Fig.~\ref{fig:distance} illustrates distances across categories for the \textbf{\textit{Cityscapes-to-FoggyCityscapes}} task, with foreground features extracted from the \textit{Source Only} \cite{ren2015faster}, \textit{AT (Baseline)} \cite{li2022cross}, and our model (\textit{Ours}). Compared to the \textit{Source Only} approach, both \textit{Baseline} and \textit{Ours} substantially reduce distances across all categories, demonstrating that adaptation strategies are essential in reducing the feature discrepancy between the source and the target domain.
Additionally, \textit{Ours} leverages the prototype cross entropy loss and mutual regularization strategy to further encourage the detector to produce more compact feature representations, resulting in a smaller $\mathcal{A}$-distance compared to the \textit{Baseline}.

\textbf{Error analysis of detection results.}
\begin{table}[!t]
\caption{TIDE error analysis. The $\Delta$$AP^{box}\texttt{@}0.5$ metric is defined as how much $AP_{50}$ can be added to the detector if an oracle fixes a certain error type in TIDE~\cite{bolya2020tide}.}
\centering
% \begin{tabular}{cc|cccccc}
\begin{tabular}{c|p{0.6cm}<{\centering}p{0.6cm}<{\centering}p{0.6cm}<{\centering}p{0.6cm}<{\centering}p{0.6cm}<{\centering}p{0.6cm}<{\centering}}
\hline
\multirow{2}{*}{Method} & \multicolumn{6}{c}{$\Delta$$AP^{box}\texttt{@}0.5$}\\ \cline{2-7}
 & \textit{cls}$\downarrow$  & \textit{loc}$\downarrow$  & \textit{both}$\downarrow$ & \textit{dup}$\downarrow$  & \textit{bg}$\downarrow$& \textit{miss}$\downarrow$  \\ \hline \hline
\textit{Source Only}  & 4.09          & \textbf{6.68} & \textbf{0.55} & 0.18          & \textbf{0.65} & 29.95 \\
\textit{AT}             & 5.02          & 10.68         & 0.69          & 0.17          & 0.86          & 19.16 \\
\textit{Ours}            & \textbf{3.89} & 11.45         & 0.84          & \textbf{0.16} & 1.08          & \textbf{16.11} \\ \hline
\end{tabular}
\label{table:error}
\end{table}
To further substantiate the effectiveness of our proposed framework for cross-domain object detection, we analyze the detection errors of the \textit{Source Only}, \textit{AT (Baseline)}, and \textit{Ours} models using the TIDE toolbox~\cite{bolya2020tide} in the context of the \textbf{\textit{Cityscapes-to-FoggyCityscapes}} task. As shown in Table~\ref{table:error}, the TIDE toolbox categorizes detection errors into six types: \emph{cls} (correct localization, incorrect classification), \emph{loc} (correct classification, incorrect localization), \emph{both} (incorrect classification and localization), \emph{dup} (multiple detections corresponding to a single ground-truth box), \emph{bg} (mistakenly classifying background as foreground), and \emph{miss} (failure to detect foreground objects).
Refer to~\cite{bolya2020tide} for comprehensive details and discussions on these error types.

We observe that the errors of \textit{Source Only} are predominantly concentrated in the \emph{miss}, with fewer other errors. It indicates that directly applying knowledge learned from the source domain to target data without adaptation leads to significant omission of objects.
Compared to \textit{Source Only}, \textit{AT (Baseline)} combines adversarial training with self-training adaptation strategies, facilitating the gradual transfer of detector knowledge to the target domain and significantly mitigating the \emph{miss} error. 
Finally, \textit{Ours} leverages the prototype cross entropy loss to encourage RoI features to be closer to the prototypes belonging to the same class, while staying far from prototypes of other classes, resulting in more compact intra-class features.
Benefiting from superior representations, \textit{Ours} performs the best in terms of the \emph{cls} and \emph{miss} error category among the three models.

\subsection{Qualitative Results}
We present more qualitative comparisons among (a) \textit{Source Only}, (b) \textit{Baseline} \cite{li2022cross}, (c) \textit{Ours}, and (d) \textit{Ground-truth} in Fig.~\ref{fig:results}. Our method can eliminate some missing errors and avoid some wrong classification cases compared with the \textit{Baseline}, which verifies the effectiveness of proposed 
PACF framework which consists of the prototype cross entropy loss and mutual regularization strategy.

\section{Conclusion}

In this paper, we propose Prototype Augmented Compact Features (PACF) framework, a simple yet effective method for domain adaptive object detection. To address the issue of large intra-class variance caused by the domain gap, we propose prototype cross entropy loss in our PACF framework, which achieves superior instance-level feature alignment. We further apply mutual regularization strategy to promote reciprocal learning between the linear and prototype-based classifiers, balancing feature compactness and discriminability.
We conduct extensive experiments under three adaptation settings, achieving state-of-the-art performance, highlighting the effectiveness of our methods.

\bibliographystyle{IEEEtran}
\bibliography{ref}

\vfill

\end{document}